\documentclass[11pt]{article}

\usepackage[T1]{fontenc}
\usepackage{newtxtext,newtxmath}

\usepackage[letterpaper,left=1.1in,right=1.1in,top=1.15in,bottom=1.15in]{geometry}
\usepackage{microtype}
\usepackage{setspace}
\usepackage{fancyhdr}
\newcommand{\shorttitle}{Radiomic Bottleneck Model for Thyroid Nodule Classification}
\fancypagestyle{plain}{%
  \fancyhf{}\fancyfoot[C]{\small\thepage}%
  }

\usepackage{titlesec}
\titleformat{\section}{\large\bfseries}{\thesection}{0.7em}{}
\titleformat{\subsection}{\normalsize\bfseries}{\thesubsection}{0.6em}{}
\titlespacing*{\section}{0pt}{2.4ex plus 0.6ex}{1.3ex}
\titlespacing*{\subsection}{0pt}{2.0ex plus 0.4ex}{0.9ex}

\usepackage[font=small,labelfont=bf,labelsep=period,skip=7pt]{caption}

\usepackage{amsmath,graphicx,booktabs,multirow}
\usepackage{tabularx}
\usepackage[backend=biber,bibstyle=ieee,citestyle=authoryear,natbib=true,doi=true,isbn=false]{biblatex}
\DeclareDelimFormat{nameyeardelim}{\addcomma\space} 
\AtBeginDocument{\let\cite\parencite} 
\AtEveryBibitem{\iffieldundef{doi}{}{\clearfield{url}}} 
\usepackage[hidelinks]{hyperref}

\title{Learning to Measure and Diagnose: A Clinical-Dictionary Radiomic Bottleneck Model for Thyroid Nodule Classification}
\author{Guibin Zhang and Weier Luo and Lei Cao}

\makeatletter
\renewcommand{\maketitle}{%
  \thispagestyle{plain}%
  \begin{center}%
    {\LARGE\bfseries\@title\par}
    \vspace{1.5em}%
    {\large Guibin Zhang$^{\dagger}$\hspace{1.2em}Weier Luo$^{\dagger}$\hspace{1.2em}Lei Cao$^{*}$\par}
    \vspace{0.6em}%
    {\normalsize\itshape School of Biomedical Engineering, Southern Medical University, Guangzhou, China\par}
    \vspace{0.5em}%
    {\small $^{\dagger}$Equal contribution.\quad $^{*}$Corresponding author: \texttt{caolei@smu.edu.cn}\par}
  \end{center}%
  \vspace{0.5em}%
}
\makeatother

\renewenvironment{abstract}
  {\par\addvspace{0.8em}\begin{center}\begin{minipage}{0.92\textwidth}\small
   \noindent\textbf{Abstract}\quad\ignorespaces}
  {\par\end{minipage}\end{center}\addvspace{0.4em}}
\newcommand{\keywords}[1]{%
  \begin{center}\begin{minipage}{0.92\textwidth}\small
  \noindent\textbf{Keywords:}\ #1\par
  \end{minipage}\end{center}\addvspace{0.4em}}

\begin{document}
\maketitle

\begin{abstract}
Thyroid ultrasound supports the assessment of nodule malignancy, but classification predictions often lack evidence linked to familiar clinical descriptions. Radiomics provides explicitly defined measurements that can support such explanations through a clinical dictionary. However, using these measurements as interpretable evidence requires reliable estimation from ultrasound images and preservation of measurement fidelity when optimizing classification performance. We propose a clinical-dictionary radiomic bottleneck model for thyroid nodule classification. Within this model, Family-Aware Radiomic Learning (FARL) estimates measurements with balanced supervision across feature families. Fidelity-Guided Bottleneck Modeling (FGBM) constructs a fixed bottleneck, followed by joint optimization of measurement estimation and classification under fidelity guidance. Decision evidence decomposition then traces feature contributions through a linear classifier and organizes them by clinical dictionary categories. Experiments on ThyroidXL and TN3K show competitive classification performance, while case studies illustrate feature contributions for benign and malignant nodules. These findings indicate interpretable classification with inspectable quantitative evidence informed by clinical knowledge. \href{https://github.com/Lsyyh/Learning-to-Measure-and-Diagnose}{Code is available online}.
\end{abstract}

\section{Introduction}
Clinicians rely on ultrasound findings to determine which thyroid nodules warrant biopsy or follow-up and which require no further evaluation~\cite{acrtirads}. Classification models can support these decisions, but a predicted label alone does not reveal which findings support the prediction. For these predictions to be clinically interpretable, their underlying evidence should be inspectable and linked to familiar ultrasound descriptions.

Existing studies have explored interpretable concepts and visual prototypes to explain image classification decisions. Concept Bottleneck Models (CBMs) map images to human-interpretable concepts and use the estimated concept values for classification~\cite{lfcbm,pcbm,adacbm}. Prototype-based methods instead relate image regions to learned prototypes or part-based concepts, providing visual evidence for their predictions~\cite{cipl,pcmnet}. However, the resulting explanations do not necessarily align with established ultrasound descriptors, limiting how directly they can be interpreted in clinical assessment.

To connect model explanations more closely with clinical assessment, radiomics provides quantitative measurements of lesion geometry, intensity, and texture, computed within segmented regions~\cite{pyradiomics}. Beyond their direct use for classification~\cite{guerrisi2024,fang2025}, these measurements can also support clinical interpretation. A clinical dictionary links selected measurements to five familiar ultrasound descriptor categories: composition, echogenicity, shape, margin, and echogenic foci~\cite{tu10}. This association provides clinical context, while explicit measurement definitions allow predicted values to be checked against measured references. However, directly using these measurements for classification requires segmentation at inference. Radiomic measurements have also been used as additional supervision during model training~\cite{rlar}, but this does not make them the primary basis of model decisions. This motivates predicting clinically associated radiomic measurements from ultrasound images and using these predictions for classification.

Using predicted, clinically associated radiomic measurements to perform image-only classification raises two interrelated challenges. First, radiomic measurements span different families that capture complementary aspects of nodule characteristics. These families may differ in learning difficulty, so improved overall estimation accuracy does not guarantee reliable predictions across them. Second, the predicted measurements must support benign--malignant classification while remaining faithful to measured references. Classification performance and measurement fidelity are not necessarily aligned, as errors in individual measurements can affect classification differently. Reducing overall estimation error may not improve diagnostic discrimination, whereas optimizing classification alone may compromise measurement fidelity.

To address these challenges, we propose a clinical-dictionary radiomic bottleneck model for thyroid nodule classification. The model estimates quantitative measurements directly from ultrasound images and performs classification without requiring segmentation at inference. Our main contributions are threefold: \textbf{(1) Family-Aware Radiomic Learning (FARL)} balances supervision across radiomic measurement families to learn complementary quantitative information. \textbf{(2) Fidelity-Guided Bottleneck Modeling (FGBM)} builds on these estimates, using measurement fidelity to guide bottleneck construction, after which it jointly optimizes estimation and classification to limit measurement drift. \textbf{(3) Decision evidence decomposition} enables feature contributions to be traced and organized by clinical dictionary categories, with a linear classifier making decisions solely from selected measurements. Experiments on two thyroid ultrasound datasets, ThyroidXL and TN3K, demonstrate competitive classification performance, while case-level analyses illustrate feature contributions for benign and malignant nodules.

\begin{figure*}[t]
\centering
\includegraphics[width=\textwidth]{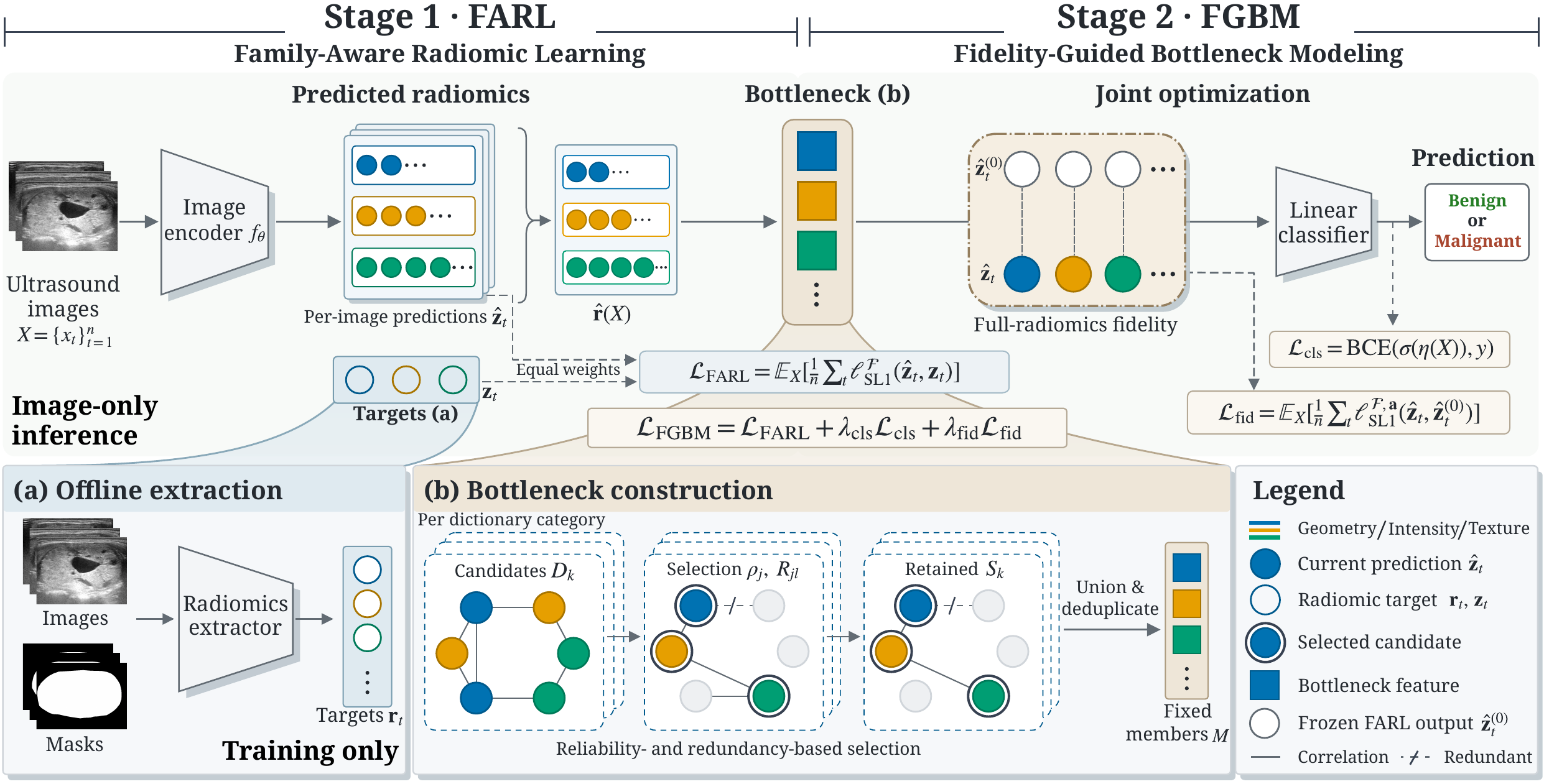}
\caption{
The proposed model is trained in two stages: Family-Aware Radiomic Learning (FARL, Stage 1) and Fidelity-Guided Bottleneck Modeling (FGBM, Stage 2). FARL learns per-image radiomic features with equally weighted family losses using offline targets from image–mask pairs, as shown in (a). Predictions are averaged per case and passed through a fixed dictionary bottleneck constructed through the reliability- and redundancy-based selection shown in (b). FGBM jointly optimizes the encoder and linear classifier with a fidelity loss anchored to frozen FARL predictions. Classification uses only bottleneck features, and inference requires ultrasound images alone.
}
\label{fig:overview}
\end{figure*}

\section{Method}
\subsection{Model Overview}
Figure~\ref{fig:overview} illustrates our model, which predicts radiomic measurements from ultrasound images and uses selected estimates as a diagnostic bottleneck. For each case, let $X=\{x_1,\ldots,x_n\}$ denote its $n$ ultrasound images and $y$ denote its label. Using the corresponding lesion masks, a reference vector $\mathbf r_t\in\mathbb{R}^{p}$ comprising $p$ radiomic features with explicit measurement definitions is extracted offline from each image $x_t$. The clinical dictionary associates selected features with clinical description categories, where $D_k$ denotes the features associated with category $k$. FARL learns to estimate the full radiomic feature vector from images, while FGBM constructs a fixed bottleneck $M\subseteq\bigcup_k D_k$ and jointly optimizes feature estimation and classification. Feature contributions from a linear classifier using only the selected predictions are traced and organized by clinical dictionary categories.

\subsection{Family-Aware Radiomic Learning}
FARL learns radiomic feature estimation from offline measurements. To reduce target skewness and scale differences, we apply a feature-wise rank-Gaussian transformation $T$ fitted on training data. The encoder $f_\theta$ predicts transformed features, which are inverse-transformed and averaged to obtain case-level estimates:
\begin{equation}
\begin{gathered}
\mathbf z_t=T(\mathbf r_t),\quad
\hat{\mathbf z}_t=f_\theta(x_t),\quad
\hat{\mathbf r}(X)=\frac{1}{n}\sum_{t=1}^{n}T^{-1}(\hat{\mathbf z}_t).
\end{gathered}
\label{eq:measure}
\end{equation}
Radiomic feature families differ in the number of features and their learning characteristics. To prevent larger families from dominating supervision, FARL averages SmoothL1 ($\ell_{\mathrm{SL1}}$) errors within each family and weights the family losses equally:
\begin{equation}
\begin{aligned}
\mathcal L_{\mathrm{FARL}}=\mathbb E_X\Bigg[
\frac{1}{n|\mathcal F|}
\sum_{t=1}^{n}\sum_{g\in\mathcal F}\
\frac{1}{|g|}\sum_{j\in g}
\ell_{\mathrm{SL1}}(\hat z_{tj},z_{tj})
\Bigg].
\end{aligned}
\label{eq:far}
\end{equation}
Here, $\mathcal F$ denotes the geometry, intensity, and texture feature families, where \(g\in\mathcal F\) indexes a family. The terms $z_{tj}$ and $\hat z_{tj}$ denote the $j$-th components of the corresponding vectors, while $\mathbb E_X$ denotes averaging over cases.

\subsection{Fidelity-Guided Bottleneck Modeling}
Building on FARL, FGBM constructs a fixed bottleneck from clinically associated radiomic features. Subsequently, it jointly optimizes feature estimation and classification under fidelity guidance.

\textbf{Bottleneck construction.} From candidates with clinical dictionary support, we retain features within each category based on reliability and redundancy. Using case-level out-of-fold predictions and corresponding reference measurements from the training data, reliability $\rho_j$ measures the Spearman correlation between predicted and reference values of feature $j$, while redundancy $R_{jl}$ measures the absolute Spearman correlation between predicted features $j$ and $l$. For each dictionary category $k$, candidates in $D_k$ are ranked by reliability. Starting with $S_k=\varnothing$, a candidate $j\in D_k$ is retained according to:
\begin{equation}
\begin{gathered}
\rho_j\geq\tau_{\mathrm{rel}},\
R_{jl}\leq\tau_{\mathrm{red}} (\forall l\in S_k) \
\Longrightarrow S_k\leftarrow S_k\cup\{j\}.
\end{gathered}
\label{eq:select}
\end{equation}
Here, $\tau_{\mathrm{rel}}$ and $\tau_{\mathrm{red}}$ denote reliability and redundancy thresholds. The final bottleneck $M=\bigcup_k S_k$ combines selected members across categories and remains fixed during subsequent optimization.

\textbf{Joint optimization.} With the selected members, FGBM jointly optimizes radiomic feature estimation and classification. For each $j\in M$, the case-level estimates are standardized and passed to a linear classifier:
\begin{equation}
\begin{gathered}
u_j(X)=\frac{\hat r_j(X)-\mu_j}{s_j},\
\eta(X)=b+\sum_{j\in M}\beta_j u_j(X).
\end{gathered}
\label{eq:head}
\end{equation}
Here, $u_j(X)$ denotes the standardized case-level estimate of feature $j$, with fixed standardization parameters $\mu_j$ and $s_j$. The coefficients $\beta_j$ and bias $b$ define the classifier, whose logit $\eta(X)$ yields the malignancy probability $\sigma(\eta(X))$.
To constrain classification-driven measurement drift, we introduce a fidelity loss relative to the frozen FARL predictions $\hat{\mathbf z}^{(0)}_t=f_{\theta_0}(x_t)$. The fixed weights $a_j$ are initialized as $1+\rho_j$ and doubled for dictionary-supported features to prioritize their fidelity preservation. They are then normalized to mean one within each radiomic family:
\begin{equation}
\begin{gathered}
\mathcal L_{\mathrm{fid}}=\mathbb E_X\Bigg[
\frac{1}{n|\mathcal F|}
\sum_{t=1}^{n}\sum_{g\in\mathcal F}\
\frac{1}{|g|}\sum_{j\in g}
a_j\ell_{\mathrm{SL1}}(\hat z_{tj},\hat z^{(0)}_{tj})
\Bigg],\\
\mathcal L_{\mathrm{FGBM}}=\mathcal L_{\mathrm{FARL}}
+\lambda_{\mathrm{cls}}\mathcal L_{\mathrm{cls}}
+\lambda_{\mathrm{fid}}\mathcal L_{\mathrm{fid}}.
\end{gathered}
\label{eq:fid_joint}
\end{equation}
Here, $\mathcal L_{\mathrm{cls}}$ is case-level binary cross-entropy, and $\lambda_{\mathrm{cls}},\lambda_{\mathrm{fid}}$ weight classification and fidelity. The measurement loss maintains supervision against reference values, while the fidelity loss constrains deviations from frozen Stage 1 predictions. Both losses cover the full radiomic feature vector to keep joint optimization anchored to the original measurement learning objective.

\subsection{Decision Evidence Decomposition}
The decision score is decomposed into contributions from selected radiomic features through the linear classifier. Let $G_j=\{k:j\in D_k\}$ denote the clinical dictionary categories associated with feature $j$. For features associated with multiple categories, their contributions are distributed equally:
\begin{equation}
\begin{gathered}
c_j(X)=\beta_j u_j(X),\quad\
C_k(X)=\sum_{j\in M:\,k\in G_j}\frac{c_j(X)}{|G_j|},\\
\eta(X)=b+\sum_k C_k(X).
\end{gathered}
\label{eq:evidence}
\end{equation}
This decomposition exactly reconstructs the decision logit, with category contributions organizing diagnostic evidence rather than representing independent predictions of clinical ultrasound findings.

\section{Experiments and Results}
\subsection{Experimental Setup}
\textbf{Dataset.} We evaluate our method on two public thyroid ultrasound datasets, ThyroidXL~\cite{thyroidxl} and TN3K~\cite{tn3k}, containing 11,554 and 3,493 image--mask pairs, respectively. Benign--malignant labels are obtained from cytology or histopathology for ThyroidXL and from TNCD~\cite{tncd} for TN3K. We use fixed training, validation, and test splits comprising 2,678/670/736 patients for ThyroidXL and 2,303/576/614 images for TN3K. Evaluation is patient-level for ThyroidXL and image-level for TN3K, per the dataset protocols.

\textbf{Implementation Details.} We use PyRadiomics to extract 102 radiomic features from native-resolution images using lesion masks, covering geometry (9), intensity (18), and texture (75). We adopt ImageNet-pretrained ConvNeXt-Tiny with $384\times384$ inputs and AdamW (learning rate $10^{-4}$, weight decay $5\times10^{-4}$, batch size 8). Training consists of 80 FARL epochs followed by 20 FGBM epochs, with $\lambda_{\mathrm{cls}}=0.5$ and $\lambda_{\mathrm{fid}}=3$. Bottleneck selection uses $\tau_{\mathrm{rel}}=0.3$ and $\tau_{\mathrm{red}}=0.9$. Checkpoints in both stages are selected by the highest mean validation $R^2$ across 102 transformed radiomic targets. Experiments are conducted in PyTorch on an NVIDIA RTX 4090 GPU and repeated with three random seeds. We report mean $\pm$ standard deviation (SD) for AUC, balanced accuracy (BACC), and the feature-averaged Spearman correlation $\bar{\rho}$ between predicted and reference values across the 102 features.

\begin{table}[t]
\centering
\caption{Comparison results (mean $\pm$ SD, \%). Bold indicates the highest mean per column within each block.}
\label{tab:comparison}
\small
\setlength{\tabcolsep}{3pt}
\begin{tabular*}{0.92\textwidth}{@{\extracolsep{\fill}}lcccc@{}}
\toprule
\multirow{2}{*}{Method} & \multicolumn{2}{c}{ThyroidXL} & \multicolumn{2}{c}{TN3K} \\
\cmidrule(lr){2-3}\cmidrule(lr){4-5}
 & AUC & BACC & AUC & BACC \\
\midrule
\multicolumn{5}{@{}l}{\textit{Radiomics references}} \\
TU1.0 (Eur. J. Radiol., 2026) & $\mathbf{99.21}$\,{\tiny$\pm0.03$} & $91.70$\,{\tiny$\pm0.18$} & $63.33$\,{\tiny$\pm1.67$} & $54.04$\,{\tiny$\pm0.52$} \\
Oracle radiomics (Cancers, 2024) & $98.90$\,{\tiny$\pm0.00$} & $\mathbf{94.72}$\,{\tiny$\pm0.00$} & $68.67$\,{\tiny$\pm0.00$} & $63.07$\,{\tiny$\pm0.00$} \\
Seg-first radiomics (JRRAS, 2025) & $87.41$\,{\tiny$\pm0.05$} & $79.68$\,{\tiny$\pm0.50$} & $\mathbf{69.84}$\,{\tiny$\pm0.12$} & $\mathbf{65.32}$\,{\tiny$\pm0.13$} \\
\midrule
\multicolumn{5}{@{}l}{\textit{Black-box references}} \\
RCAF (Sci. Rep., 2026) & $79.41$\,{\tiny$\pm3.75$} & $62.92$\,{\tiny$\pm3.76$} & $65.66$\,{\tiny$\pm3.57$} & $60.89$\,{\tiny$\pm2.27$} \\
RLAR (arXiv, 2026) & $70.86$\,{\tiny$\pm3.04$} & $62.81$\,{\tiny$\pm2.66$} & $63.08$\,{\tiny$\pm9.19$} & $60.48$\,{\tiny$\pm7.70$} \\
Direct classifier & $\mathbf{94.98}$\,{\tiny$\pm0.37$} & $\mathbf{88.06}$\,{\tiny$\pm0.79$} & $\mathbf{81.32}$\,{\tiny$\pm0.35$} & $\mathbf{73.29}$\,{\tiny$\pm1.40$} \\
\midrule
\multicolumn{5}{@{}l}{\textit{Interpretable methods}} \\
LF-CBM (ICLR, 2023) & $67.06$\,{\tiny$\pm0.53$} & $50.11$\,{\tiny$\pm0.17$} & $63.14$\,{\tiny$\pm1.02$} & $55.64$\,{\tiny$\pm1.12$} \\
PCBM (ICLR, 2023) & $62.56$\,{\tiny$\pm0.25$} & $52.00$\,{\tiny$\pm1.89$} & $60.48$\,{\tiny$\pm0.27$} & $53.94$\,{\tiny$\pm2.83$} \\
AdaCBM (MICCAI, 2024) & $85.28$\,{\tiny$\pm0.56$} & $65.27$\,{\tiny$\pm2.55$} & $69.70$\,{\tiny$\pm0.47$} & $62.06$\,{\tiny$\pm1.68$} \\
CIPL (IEEE TMI, 2025) & $94.08$\,{\tiny$\pm0.44$} & $86.02$\,{\tiny$\pm1.19$} & $72.06$\,{\tiny$\pm2.23$} & $65.97$\,{\tiny$\pm2.10$} \\
PCMNet (AAAI, 2026) & $94.02$\,{\tiny$\pm0.33$} & $75.68$\,{\tiny$\pm3.18$} & $79.27$\,{\tiny$\pm0.95$} & $68.74$\,{\tiny$\pm2.11$} \\
\textbf{Ours} & $\mathbf{94.77}$\,{\tiny$\pm0.14$} & $\mathbf{86.73}$\,{\tiny$\pm0.55$} & $\mathbf{79.49}$\,{\tiny$\pm1.62$} & $\mathbf{72.23}$\,{\tiny$\pm0.90$} \\
\bottomrule
\end{tabular*}
\par\vspace{3pt}
\begin{minipage}{0.92\textwidth}\raggedright\small
RCAF and Seg-first radiomics use predicted masks. TU1.0 and Oracle radiomics require ground-truth masks at inference.
\end{minipage}
\end{table}

\subsection{Comparison Experiment}
Table~\ref{tab:comparison} compares our method with radiomics-based, black-box, and interpretable classification approaches. Our method achieves mean AUC and BACC of $94.77\%$ and $86.73\%$ on ThyroidXL, with $79.49\%$ and $72.23\%$ on TN3K. Among radiomics references, TU1.0 and Oracle radiomics achieve higher performance on ThyroidXL but require ground-truth masks at inference. In contrast, Seg-first radiomics uses predicted masks but performs worse than our method on both datasets, while our approach outperforms all three radiomics references on TN3K. Compared with black-box methods, our method outperforms RCAF~\cite{rcaf} and RLAR on both datasets, with AUC only 0.21 and 1.83 percentage points below direct image classification. Within interpretable methods, our approach achieves the highest mean AUC and BACC on both datasets. These results demonstrate competitive classification performance while relying solely on predicted radiomic features and providing traceable quantitative evidence.

\begin{table}[t]
\centering
\caption{Ablation results (mean $\pm$ SD; AUC in \%, $\bar{\rho}\times100$). Bold indicates the highest mean per column.}
\label{tab:ablation}
\small
\setlength{\tabcolsep}{4pt}
\begin{tabular}{p{0.30\textwidth}cccc}
\toprule
\multirow{2}{*}{Configuration} & \multicolumn{2}{c}{ThyroidXL} & \multicolumn{2}{c}{TN3K} \\
\cmidrule(lr){2-3}\cmidrule(lr){4-5}
 & AUC & $\bar{\rho}$ & AUC & $\bar{\rho}$ \\
\midrule
Full model & $\mathbf{94.77}$\,{\tiny$\pm0.14$} & $\mathbf{91.34}$\,{\tiny$\pm0.08$} & $\mathbf{79.49}$\,{\tiny$\pm1.62$} & $84.19$\,{\tiny$\pm0.55$} \\
w/o family balancing & $94.05$\,{\tiny$\pm0.33$} & $91.00$\,{\tiny$\pm0.18$} & $78.82$\,{\tiny$\pm0.55$} & $84.66$\,{\tiny$\pm0.19$} \\
w/o joint optimization & $92.20$\,{\tiny$\pm0.64$} & $91.22$\,{\tiny$\pm0.19$} & $73.24$\,{\tiny$\pm0.09$} & $\mathbf{84.90}$\,{\tiny$\pm0.18$} \\
w/o fidelity loss & $93.90$\,{\tiny$\pm0.85$} & $90.42$\,{\tiny$\pm0.50$} & $79.26$\,{\tiny$\pm1.75$} & $83.06$\,{\tiny$\pm0.65$} \\
\bottomrule
\end{tabular}
\end{table}

\subsection{Ablation Study}
Table~\ref{tab:ablation} evaluates the key designs of FARL and FGBM. As shown, omitting family balancing slightly reduces mean AUC on both datasets, whereas removing joint optimization causes larger declines of 2.57 and 6.25 percentage points on ThyroidXL and TN3K, respectively. In contrast to the consistent AUC reductions, measurement fidelity exhibits different trends across datasets. With separate training, $\bar{\rho}$ decreases slightly from 91.34 to 91.22 on ThyroidXL but increases from 84.19 to 84.90 on TN3K. This difference underscores the importance of considering measurement fidelity alongside classification performance. Moreover, without the fidelity loss, $\bar{\rho}$ decreases on both datasets, accompanied by lower AUC. These results show that fidelity guidance helps preserve the correspondence between predicted and measured radiomic features, supporting the interpretation of feature contributions.

\begin{figure}[!t]
\centering
\includegraphics[width=0.62\textwidth]{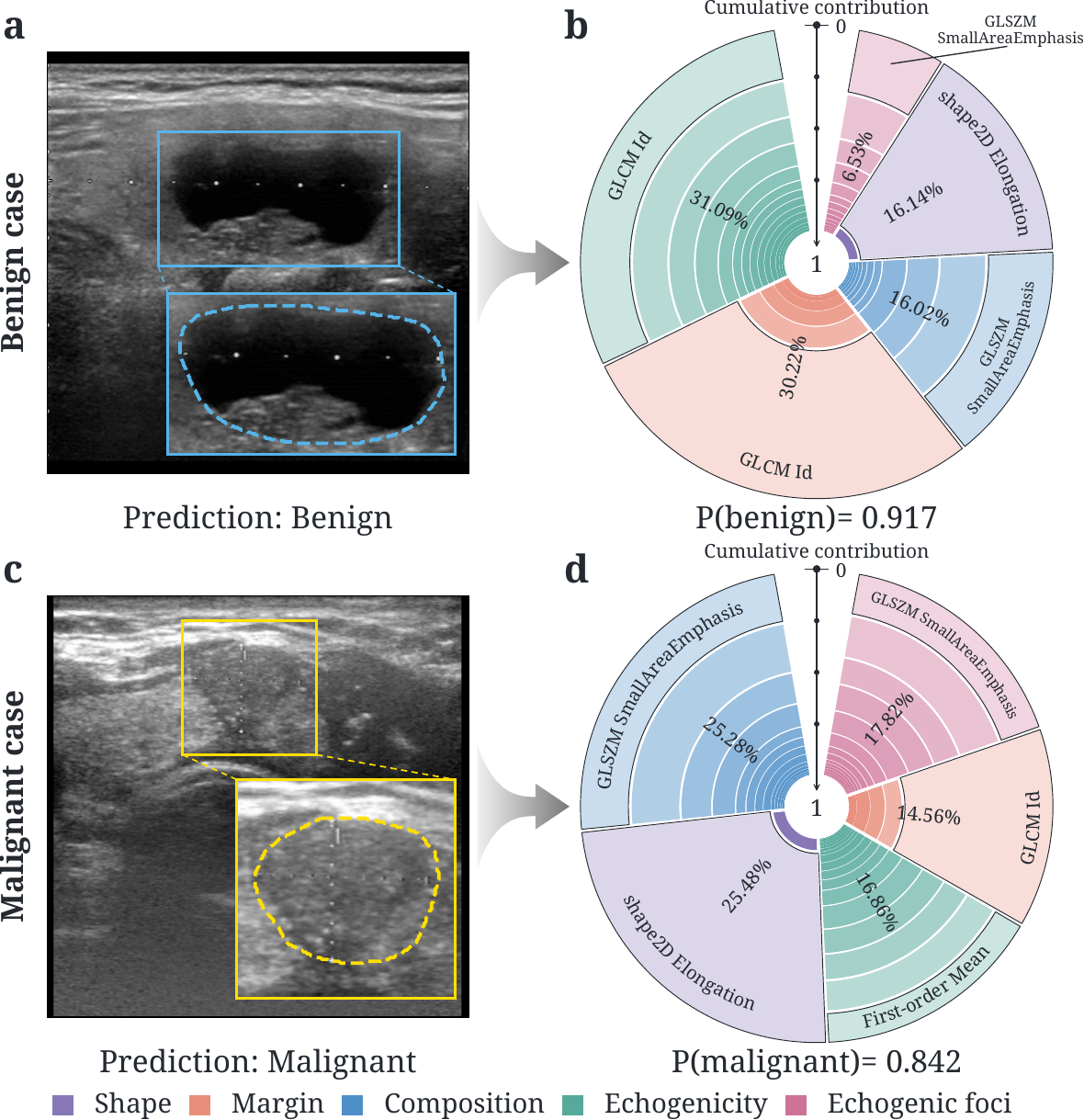}
\caption{Feature contributions for representative TN3K cases. (a, c) Ultrasound images of the benign and malignant nodules. (b, d) Corresponding contributions organized by clinical dictionary categories. Sector percentages indicate the relative magnitudes of category contributions $C_k$. Radial bands show cumulative absolute feature contributions within each category, with the dominant feature labeled.}
\label{fig:cases}
\end{figure}

\subsection{Case Study}
Figure~\ref{fig:cases} illustrates how the clinical dictionary organizes quantitative evidence from contrasting nodule appearances. Net category contributions are dominated by echogenicity and margin in the benign case, but by shape and composition in the malignant case. Within these categories, \textit{GLCM Id} reflects the benign nodule's relatively uniform texture, whereas \textit{shape2D Elongation} and \textit{GLSZM SmallAreaEmphasis} reflect the malignant nodule’s elongation and prevalence of small homogeneous zones. Taken together, these findings show how the model combines quantitative evidence and clinical context to explain individual predictions.

\section{Conclusion}
This paper presents a clinical-dictionary radiomic bottleneck model for interpretable thyroid nodule classification from ultrasound images. Specifically, FARL balances supervision across feature families, while FGBM jointly optimizes measurement estimation and classification under fidelity guidance. Decision evidence decomposition traces feature contributions and organizes them by clinical dictionary categories. Results on ThyroidXL and TN3K demonstrate a balance between classification performance and measurement fidelity. In addition, the model makes the quantitative basis of individual predictions inspectable through traceable feature contributions associated with ultrasound descriptions. Future work will extend this model to more medical scenarios by incorporating task-specific quantitative measurements and clinical knowledge.

\par\bigskip
\noindent\textbf{Compliance with Ethical Standards.}
This retrospective study used the publicly available ThyroidXL~\cite{thyroidxl} and TN3K~\cite{tn3k} datasets, with no additional participant recruitment or data collection.

\par\medskip
\noindent\textbf{Acknowledgments.}
No funding was received for conducting this study. The authors have no relevant financial or non-financial interests to disclose.
GPT-6 Sol (OpenAI; released September 22, 2026) assisted with code development and revision, organization and visualization of results, and manuscript language editing. All AI-generated content was reviewed by the authors.

\clearpage
\printbibliography

\end{document}